\pdfoutput=1
\documentclass[11pt]{article}
\usepackage[preprint]{acl}
\usepackage{times}
\usepackage{latexsym}
\usepackage[T1]{fontenc}
\usepackage[utf8]{inputenc}
\usepackage{microtype}
\usepackage{inconsolata}
\usepackage{graphicx}
\usepackage{amsmath,amssymb}
\usepackage{booktabs}
\usepackage{multirow}
\usepackage{xcolor}
\usepackage{bm}
\usepackage{placeins}

\newcommand{\act}[1]{\texttt{#1}}
\definecolor{oursbg}{RGB}{253,240,228}
\newcommand{\ours}{\cellcolor{oursbg}}
\usepackage[table]{xcolor}

\title{Why Generate a Click? Plug-in Decision Heads for Web Agents}

\author{
  \textbf{Chengguang Gan\textsuperscript{1}},
  \textbf{Yunhao Liang\textsuperscript{2}},
  \textbf{Hanjun Wei\textsuperscript{2}},
\\
  \textbf{Qinghao Zhang\textsuperscript{3}},
  \textbf{Zhixi Cai\textsuperscript{4}},
  \textbf{Shiwen Ni\textsuperscript{5}}
\\
  \textsuperscript{1}Independent Researcher, \quad
  \textsuperscript{2}University of Chinese Academy of Sciences,
\\
  \textsuperscript{3}Pusan National University, \quad
  \textsuperscript{4}Monash University,
\\
  \textsuperscript{5}Shenzhen University of Advanced Technology
\\[4pt]
  \small{\textbf{Correspondence:} \href{mailto:chengguangg1024@gmail.com}{chengguangg1024@gmail.com}}
}

\begin{document}
\maketitle

\begin{abstract}
Web agents built on vision language models decide every action by generating text such as \act{click(12)}, although most of that text encodes a choice among a few action types and a few dozen marked elements. We read these choices directly from the hidden states that the generator computes while it reads the goal and the screenshot. Plug-in decision heads on a frozen generator predict the action type, pool the visual tokens inside each numbered box to pick the target, and point at a listed text option or at a WRITE slot that leaves the writing of new text to the generator. A confidence cascade executes the heads' action when they are sure and otherwise lets the generator decode from the KV cache of the same forward pass. With Qwen3-VL 4B and 8B generators trained on 213k web steps, a strict threshold keeps the cascade between 0.3 points below and 1.4 points above the generator on seven offline test sets while the heads decide 12 to 64\% of the steps. A looser threshold raises step success on unseen Mind2Web websites by 3.1 and 1.8 points, costs up to 2 points on GUIAct and MiniWoB++, and lifts WebArena task success from 12.5 to 13.8\% (4B) and from 13.6 to 14.6\% (8B) with less than half of the generated tokens. Training the heads jointly with the generator lowers the generator's own step success by up to 4.9 points; heads on a frozen generator avoid this cost.
\end{abstract}

\section{Introduction}

\begin{figure}[t]
  \centering
  \includegraphics[width=\columnwidth]{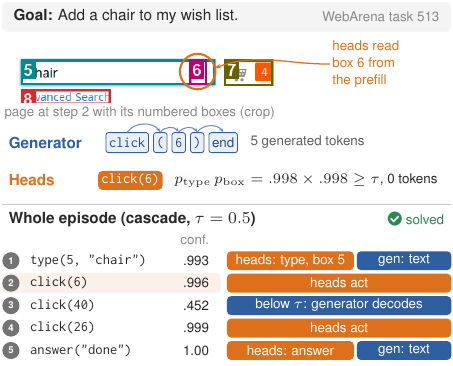}
  \caption{WebArena task 513 as solved by our 4B cascade ($\tau=0.5$, seed 0). On the page of step 2 the generator would spell \act{click(6)} as five generated tokens, while the heads read the type and the box from the generator's prefill and act without generating a token. Over the episode the heads act alone twice, twice choose the action while the generator writes only the text, and hand one uncertain step to the generator (more episodes in Appendix~\ref{app:cases}).}
  \label{fig:example}
\end{figure}

\begin{figure*}[t]
  \centering
  \includegraphics[width=\textwidth]{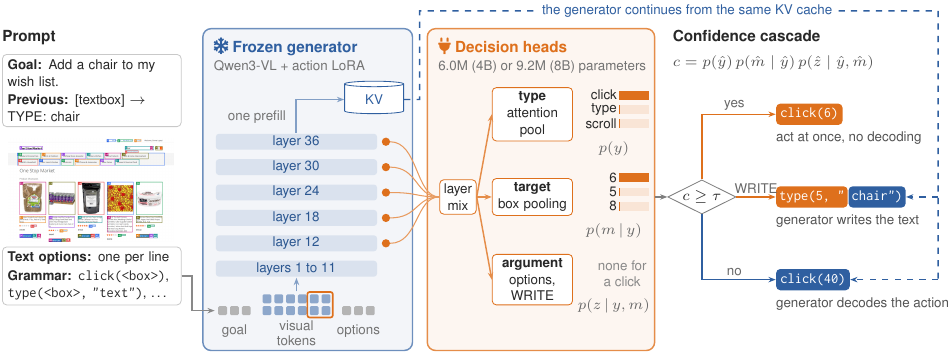}
  \caption{Overview. The frozen generator reads the goal, the history, the marked screenshot and the listed text options in one prefill. Decision heads read a learned mix of five of its layers and return a distribution over action types, boxes and options (with a WRITE slot). When the confidence $c_t$ reaches $\tau$ the agent acts at once or, for WRITE, lets the generator write only the text; otherwise the generator decodes the whole action from the same KV cache.}
  \label{fig:overview}
\end{figure*}

A web agent built on a vision language model acts by writing. At each step the model reads the goal and a screenshot whose elements carry numbered boxes, then generates a string such as \act{click(6)} or \act{type(5, "chair")} that a harness parses and executes \citep{zheng2024gpt,koh2024visualwebarena,xu2024aguvis,qin2025ui}. Most of this string records a choice from a short list: one of eleven action types and one of a few dozen boxes. Only the text of a typing or answering step is open ended. Writing a choice out token by token costs one sequential decoding pass per token, and it scores only the one string written, with no explicit distribution over whole alternative actions.

By the time the generator writes its first token, it has read the goal, the history and every box on the screen, and that first token already commits to an action type. Like other properties that probes recover from intermediate layers \citep{alain2016understanding}, and like the knowledge that probing and editing studies locate inside large language and multimodal models \citep{jiang2026mined,jiang2025mmke,qi2025context}, the decision should therefore be readable from the hidden states of the prefill (Figure~\ref{fig:example}). This suggests a cheaper way to act: read the decision from those states with small heads, and generate only what the heads cannot express. Two questions determine whether this is worth doing. On which steps do heads read the decision as reliably as the generator writes it? And can an agent use the heads on those steps and generation on the rest without paying for both?

Our answer is a set of plug-in decision heads (Figure~\ref{fig:overview}). A type head pools the whole prefill, a target head pools the visual tokens inside each numbered box and scores the box against a learned query, and an argument head points at the text options listed in the prompt or at a WRITE slot that returns the writing of new text to the generator. The heads read a learned mix of five layers of a frozen generator and add about 0.1\% to its parameters. A confidence cascade executes the heads' action when the product of their probabilities reaches a threshold $\tau$. Otherwise the generator decodes from the KV cache of the forward pass the heads already read.

We train Qwen3-VL generators of 4B and 8B parameters on 213k web steps from MiniWoB++, Mind2Web, GUIAct, UGround and WebLINX, and then train heads on each frozen generator. At $\tau=0.9$ the cascade matches its generator on all seven offline test sets, from 0.3 points below to 1.4 above, while the heads decide 12 to 64\% of the steps. At $\tau=0.5$ step success on Mind2Web websites unseen in training rises by 3.1 (4B) and 1.8 (8B) points, and the four GUIAct test sets lose 1.0 to 1.9 points. On WebArena the cascade completes 13.8\% and 14.6\% of 174 test tasks, against 12.5\% and 13.6\% for the generators, and generates 3.1 and 3.3 tokens per step instead of 7.4 and 7.5; the gain comes from tasks whose intent template also occurs in training. Training the heads together with the generator in one adapter lowers the generator's own step success by 0.5 to 4.9 points, and on Mind2Web and WebArena this cost outweighs the stronger heads that joint training yields.

Decision heads give an existing web agent a second and cheaper way to act without retraining it. The generator stays untouched, so every step the heads hand back is decided exactly as before, and every step they decide comes with an explicit distribution over types, boxes and options that a harness can threshold, log or search over \citep{koh2024tree}. Decoding is then reserved for uncertain steps and for text that the option list does not contain.

\section{Related Work}

\paragraph{Web agents that act by generation.}
MiniWoB++ \citep{shi2017world,liu2018reinforcement}, Mind2Web \citep{deng2023mind2web}, WebArena \citep{zhou2024webarena} and VisualWebArena \citep{koh2024visualwebarena} evaluate agents that map a page and a goal to an action. Current vision language agents write this action as text, either as element ids on Set-of-Mark screenshots \citep{yang2023set,zheng2024gpt} or as screen coordinates after large-scale grounding training \citep{cheng2024seeclick,hong2024cogagent,gou2025navigating,wu2025atlas,xu2024aguvis,qin2025ui}, with parsers such as OmniParser \citep{lu2024omniparser} supplying the elements. Earlier agents used dedicated output layers: CC-Net \citep{humphreys2022data} predicts factored actions with policy heads, MindAct \citep{deng2023mind2web} ranks candidate elements before a language model chooses among them, and GUI-Actor \citep{wu2026gui} replaces coordinate generation with an attention head over image patches, which can also be trained on a frozen backbone.

\paragraph{Spending less computation per decision.}
Early exit methods stop at an intermediate layer once a classifier is confident \citep{teerapittayanon2016branchynet,xin2020deebert,schuster2022confident}, selective classifiers abstain below a confidence threshold \citep{geifman2017selective}, cascades send easy queries to cheaper models \citep{chen2023frugalgpt}, speculative decoding and Medusa draft tokens that the full model verifies \citep{leviathan2023fast,cai2024medusa}, and visual token compression shortens the multimodal prefill itself \citep{peng2025can}. SwiftSage pairs a fast small agent with a slow large one \citep{lin2023swiftsage}. These methods save computation within text generation or across models, and the agent's decision is still written out as text. Appendix~\ref{app:related} discusses further related work.

\paragraph{What is new.}
Earlier action heads either replace the language model, as in CC-Net, or replace generation for one part of the action, as in MindAct and GUI-Actor. Our heads read the whole action, its type, target box and text option, from the prefill of a frozen generator agent and return control to that generator through a WRITE slot or a confidence threshold, so heads and generation become two readouts of one forward pass. We also measure what joint training costs the agent's own generation. Decision heads thus become an add-on for existing generator agents.

\section{Plug-in Decision Heads}

\subsection{Observation and Action Space}
\label{sec:obs}

An episode comes with a goal $g$. At step $t$ the agent sees a screenshot $I_t$ on which each of $N$ candidate elements is drawn as a numbered box, the history $h_t$ of its earlier actions, one line each in Mind2Web's text form with the element's role and visible text (for example \act{[button] Search -> CLICK}), and a list $O_t$ of text options, one per line: strings taken from the goal and the page and the choices of select elements, listed without the element they belong to (Appendix~\ref{app:data}). The prompt $x_t$ places $g$ and $h_t$ before the image, so that under causal attention every visual token is computed with the task already in view, and puts $O_t$ and the action grammar after it (Figure~\ref{fig:overview}, left).

An action is a triple $a=(y,m,z)$ of a type $y\in\mathcal{Y}$, a target box $m\in\{1,\dots,N\}$ and an argument $z$. The eleven types are click, type, select, hover, clear, enter, scroll down, scroll up, go back, answer and no operation; the last five take no box. Typing needs a text, selecting needs one of the element's choices, and answering needs the answer text. The generator $G_\theta$, a vision language model with a LoRA adapter \citep{hu2021lora}, writes $a$ as a string $w$ such as \act{type(5, "chair")} and learns from demonstrations with
\begin{equation}
\mathcal{L}_{\mathrm{gen}}(\theta)=-\sum_{k=1}^{|w|}\log p_\theta\left(w_k\mid w_{<k},x_t\right).
\label{eq:gen}
\end{equation}

\subsection{Heads on the Generator's Prefill}
\label{sec:heads}

Before it writes $w_1$, the generator runs one forward pass over $x_t$. This prefill yields hidden states $H^{(\ell)}\in\mathbb{R}^{T\times d}$ at every layer $\ell$, the KV cache that decoding continues from, and the logits of the first token. The heads read a learned mix of five of the 36 layers, $\mathcal{S}=\{12,18,24,30,36\}$, in the manner of a scalar mix \citep{peters-etal-2018-deep}, followed by layer normalization (LN):
\begin{equation}
\tilde H=\mathrm{LN}\Big(\sum_{\ell\in\mathcal{S}}\beta_\ell H^{(\ell)}\Big),\qquad \beta=\mathrm{softmax}(\alpha).
\label{eq:mix}
\end{equation}
Three learned queries $q_c\in\mathbb{R}^{d_h}$, one for each head $c\in\{\mathrm{type},\mathrm{tgt},\mathrm{arg}\}$, summarize the whole sequence by attention pooling:
\begin{equation}
u_c=\sum_{i=1}^{T}\mathrm{softmax}_i\Big(\frac{q_c^{\top}W_K\tilde h_i}{\sqrt{d_h}}\Big)\,W_V\tilde h_i .
\label{eq:pool}
\end{equation}

\paragraph{Type head.}
$p(y\mid x_t)=\mathrm{softmax}\big(\mathrm{MLP}(u_{\mathrm{type}})\big)$, restricted to the types the page allows: typing needs a text field among the boxes, selecting an element with choices, and hovering a menu item. The harness reads these element roles from the page.

\paragraph{Target head.}
Qwen3-VL \citep{bai2025qwen3} resizes the screenshot to multiples of 32 pixels and encodes each $32\times32$ cell as one visual token. We map every box into the resized image, and $P_m$ holds the positions of the cells that box $m$ touches; a box smaller than a cell receives the cell that contains it. The head averages these tokens and scores the box against the target summary:
\begin{equation}
v_m=\frac{1}{|P_m|}\sum_{i\in P_m}\tilde h_i,\quad s_m=\frac{(W_B v_m)^{\top}W_Q u_{\mathrm{tgt}}}{\sqrt{d_h}}
\label{eq:box}
\end{equation}
and $p(m\mid y,x_t)$ is the softmax of $s_m$ over the boxes that can take type $y$ (every box for a click, text fields for typing). Each box is represented by what the generator computed for its pixels, so the target head needs the box coordinates and no element text.

\paragraph{Argument head.}
Let $r_k$ be the hidden state at the last token of option line $k$. In the manner of a pointer network \citep{vinyals2015pointer}, the head scores every option and a learned key $r_{\mathrm{W}}$ for WRITE:
\begin{equation}
o_k=\frac{(W_A r_k)^{\top}W_{Q'}u_{\mathrm{arg}}}{\sqrt{d_h}},\quad o_{\mathrm{W}}=\frac{r_{\mathrm{W}}^{\top}W_{Q'}u_{\mathrm{arg}}}{\sqrt{d_h}}
\label{eq:arg}
\end{equation}
For typing, $p(z\mid y,m,x_t)$ is the softmax over the text options and WRITE; for selecting, over the choices of box $m$. WRITE states that the text to type is not among the options, so the generator has to write it. Both scores condition on $y$ and $m$ through the set of boxes or options that each softmax ranges over.

With $d_h=512$ the heads hold 6.0M parameters on the 4B generator and 9.2M on the 8B one, about 0.1\% of the backbone.

\subsection{Training on a Frozen Generator}
\label{sec:train}

We first train the generator alone with Eq.~\ref{eq:gen}. We then freeze $\theta$ and train only the heads' parameters $\phi$ on the same demonstrations:
\begin{equation}
\begin{split}
\mathcal{L}_{\mathrm{heads}}(\phi)=&-\log p(y^{\star})-\log p(m^{\star}\mid y^{\star})\\
&-\log p(z^{\star}\mid y^{\star},m^{\star})+0.3\,\mathcal{L}_{\mathrm{done}},
\end{split}
\label{eq:heads}
\end{equation}
where $(y^{\star},m^{\star},z^{\star})$ is the demonstrated action, $z^{\star}$ is WRITE whenever the demonstrated text is not a listed option, the target and argument terms vanish for types without them, and $\mathcal{L}_{\mathrm{done}}$ is an auxiliary binary cross entropy, from a linear output on $u_{\mathrm{type}}$, on whether the step completes the task, used only as a training signal. Since $\theta$ stays fixed, the generator's own actions do not change.

\subsection{Confidence Cascade with a Shared Prefill}
\label{sec:cascade}

The heads decide greedily,
\begin{equation}
\begin{split}
\hat y&=\textstyle\arg\max_{y}\,p(y\mid x_t),\\
\hat m&=\textstyle\arg\max_{m}\,p(m\mid \hat y,x_t),\\
\hat z&=\textstyle\arg\max_{z}\,p(z\mid \hat y,\hat m,x_t),
\end{split}
\end{equation}
and report the action confidence
\begin{equation}
c_t=p(\hat y\mid x_t)\,p(\hat m\mid \hat y,x_t)\,p(\hat z\mid \hat y,\hat m,x_t),
\label{eq:conf}
\end{equation}
where the factor of an absent part is 1. With a threshold $\tau$ and W denoting the WRITE slot, the agent takes
\begin{equation}
a_t=\begin{cases}
(\hat y,\hat m,\hat z) & c_t\ge\tau,\ \hat z\ne\mathrm{W},\\
(\hat y,\hat m,G_\theta(\cdot\mid x_t,\pi_t)) & c_t\ge\tau,\ \hat z=\mathrm{W},\\
G_\theta(\cdot\mid x_t) & c_t<\tau,
\end{cases}
\label{eq:cascade}
\end{equation}
where $\pi_t$ is the start of the action string, such as \act{type(5, "}. We say the heads \emph{decide} a step when $c_t\ge\tau$ and \emph{act alone} when, in addition, $\hat z\ne\mathrm{W}$.

A step handed back starts decoding from the cached prefill of Section~\ref{sec:heads}. For WRITE, the harness appends $\pi_t$ to the cached prompt in one forward pass, the generator decodes at most 24 tokens, and the text up to the first quote becomes the argument. An answer always takes this path with $\pi_t=$ \act{answer("}, and its confidence is $p(\mathrm{answer}\mid x_t)$. A handed back step therefore costs what the generator alone would spend on it plus a few matrix products, and a step on which the heads act alone costs one prefill and no decoding. A threshold above 1 recovers the generator exactly.

\section{Experimental Setup}

\begin{table*}[t]
\centering
\small
\setlength{\tabcolsep}{6pt}
\begin{tabular}{llccccccc}
\toprule
& & \multicolumn{3}{c}{M2W} & \multicolumn{4}{c}{GUIAct web} \\
\cmidrule(lr){3-5}\cmidrule(lr){6-9}
& System & Task & Website & Domain & Single & Multi & Single (ext.) & Multi (ext.) \\
\midrule
\multirow{6}{*}{4B} & Generator & 46.2 & 40.9 & 44.3 & \textbf{84.1} & \textbf{56.1} & \textbf{78.9} & 63.5 \\
& Heads only & 39.7 & 37.9 & 39.1 & 81.0 & 54.7 & 75.8 & 60.3 \\
& \ours Cascade, $\tau=0.9$ & \ours 46.6 & \ours 42.3 & \ours 44.4 & \ours 84.0 & \ours 55.9 & \ours 78.8 & \ours \textbf{63.7} \\
& \ours Cascade, $\tau=0.5$ & \ours \textbf{47.6} & \ours \textbf{44.0} & \ours \textbf{44.6} & \ours 82.9 & \ours 54.2 & \ours 77.9 & \ours 61.8 \\
& Joint, generation & 42.0 & 36.0 & 42.4 & 83.6 & 52.1 & 78.1 & 60.5 \\
& Joint, cascade $\tau=0.5$ & 45.5 & 40.8 & 44.5 & 83.2 & 54.3 & 77.7 & 61.9 \\
\midrule
\multirow{6}{*}{8B} & Generator & 50.4 & 44.5 & 48.1 & \textbf{84.4} & \textbf{58.8} & \textbf{79.3} & \textbf{64.1} \\
& Heads only & 41.7 & 41.2 & 41.7 & 81.3 & 54.9 & 76.1 & 60.3 \\
& \ours Cascade, $\tau=0.9$ & \ours \textbf{50.6} & \ours 45.9 & \ours 48.1 & \ours 84.1 & \ours 58.7 & \ours 79.0 & \ours 64.0 \\
& \ours Cascade, $\tau=0.5$ & \ours 50.2 & \ours 46.3 & \ours \textbf{48.2} & \ours 82.8 & \ours 56.9 & \ours 78.0 & \ours 62.6 \\
& Joint, generation & 46.1 & 42.1 & 46.2 & 83.6 & 54.9 & 78.6 & 61.4 \\
& Joint, cascade $\tau=0.5$ & 48.7 & \textbf{46.5} & 47.4 & 82.9 & 56.4 & 77.8 & 61.8 \\
\bottomrule
\end{tabular}
\caption{Offline step success (\%). Our cascade (shaded) uses heads trained on the frozen generator of the first row. The heads decide 12 to 18\% of the M2W steps and 27 to 64\% of the GUIAct steps at $\tau=0.9$, and 37 to 46\% and 66 to 89\% at $\tau=0.5$.}
\label{tab:offline}
\end{table*}

\subsection{Data and Backbones}
\label{sec:data}

The generators are Qwen3-VL-4B-Instruct and Qwen3-VL-8B-Instruct \citep{bai2025qwen3} with a rank 16 LoRA adapter on every attention and MLP projection of the language model. They train for one epoch on 213k steps that we convert into the format of Section~\ref{sec:obs}: successful MiniWoB++ episodes on training seeds (9.2k steps, repeated three times), the Multimodal-Mind2Web training split \citep{deng2023mind2web,zheng2024gpt} (7.3k), GUIAct web single and multi step tasks \citep{chen2025guicourse} with scroll and answer steps (94k), web grounding data from UGround \citep{gou2025navigating} (65k) and WebLINX demonstrations \citep{lu2024weblinx} (38k). The heads train for one epoch on the same steps with the generator frozen. Appendix~\ref{app:data} describes the conversion and Appendix~\ref{app:train} the optimization.

For WebArena, the generator and then its heads continue for three epochs on steps that we replay in the live sites from successful expert trajectories on 115 training tasks disjoint from the test tasks (Appendix~\ref{app:wa}).

\subsection{Benchmarks and Metrics}
\label{sec:bench}

\paragraph{Offline.}
We use the three Multimodal-Mind2Web (M2W) test splits, cross-task, cross-website and cross-domain, whose tasks, websites or domains are unseen in training, and the GUIAct web test sets for single and multi step tasks together with their extended versions (ext.) that add scroll and answer steps. For each M2W step we cut a $1280\times1000$ window at a random height that keeps the gold element visible and draw up to 64 candidates: the gold element, the other positive candidates and random negatives. GUIAct steps likewise show the target and up to 63 other elements. This gives 1,253, 973 and 3,820 M2W steps and 1,089, 421, 1,410 and 854 GUIAct steps; Section~\ref{sec:std} also counts, as failures, the 372 M2W steps whose gold element cannot be shown. A step succeeds when the chosen box is a gold box and the operation, including any text, matches.

\paragraph{Online.}
MiniWoB++ runs all 119 tasks with five seeds each through BrowserGym \citep{chezelles2024browsergym}, at most 10 steps per episode; 12 tasks are held out from training. WebArena \citep{zhou2024webarena} runs 174 test tasks on six sites with at most 25 steps per task. Success follows the WebArena checkers, with GPT-5.5 judging answers that need fuzzy matching, and every run starts from freshly reset sites.

Every success rate we report is the mean over three training seeds (per-seed results in Appendix~\ref{app:seeds}).

\subsection{Compared Systems}
\label{sec:systems}

\textbf{Generator}: the LoRA generator alone, decoding greedily without masks; an output that does not parse is an invalid action. \textbf{Heads only}: our heads act on every step; a WRITE falls back to the best listed text and an answer is left empty. \textbf{Cascade}: Eq.~\ref{eq:cascade} with the frozen generator, at $\tau=0.9$ and $\tau=0.5$. We fixed both thresholds in early MiniWoB++ experiments: 0.9 as a strict setting and 0.5, at which the heads decide only when their action holds at least half of the probability. \textbf{Joint}: one adapter trained on $\mathcal{L}_{\mathrm{gen}}+\mathcal{L}_{\mathrm{heads}}$ with equal weights, the same data and the same schedule, used for generation alone or in the same cascade. All systems share the prompt and the boxes. Section~\ref{sec:std} compares with published M2W results, and Section~\ref{sec:analysis} adds two-model cascades, a patch attention target head and a generator that writes a thought before each action.

\section{Results}

\subsection{Offline Benchmarks}
\label{sec:offline}

Table~\ref{tab:offline} compares the systems on the seven offline test sets. At $\tau=0.9$ the cascade tracks its generator on every set while the heads decide between 12\% (M2W cross-website) and 64\% (GUIAct single, ext.) of the steps. An agent can therefore run at $\tau=0.9$ by default.

The looser threshold $\tau=0.5$ moves accuracy in opposite directions on the two benchmarks. On M2W it improves every split for the 4B model, by up to 3.1 points on unseen websites, and the cross-website split for the 8B model by 1.8. On GUIAct it costs 1.0 to 1.9 points at both scales. Section~\ref{sec:where} locates the steps responsible.

Heads only trails the generator by 1.4 to 8.7 points; the threshold confines the heads to the steps they read well. The joint rows start from a weaker generator, which Section~\ref{sec:joint} examines.

\subsection{Online Agents on MiniWoB++ and WebArena}
\label{sec:online}

\begin{table}[t]
\centering
\small
\setlength{\tabcolsep}{3.2pt}
\begin{tabular}{lcccccc}
\toprule
& \multicolumn{2}{c}{MiniWoB++} & \multicolumn{2}{c}{WebArena} & \multicolumn{2}{c}{Tokens/step} \\
\cmidrule(lr){2-3}\cmidrule(lr){4-5}\cmidrule(lr){6-7}
System & 4B & 8B & 4B & 8B & 4B & 8B \\
\midrule
Generator & \textbf{53.3} & \textbf{55.5} & 12.5 & 13.6 & 7.4 & 7.5 \\
\ours Cascade, $\tau=0.5$ & \ours 52.5 & \ours 53.6 & \ours \textbf{13.8} & \ours \textbf{14.6} & \ours 3.1 & \ours 3.3 \\
Joint, generation & \textbf{53.3} & 55.0 & 11.5 & 11.1 & 7.3 & 7.4 \\
Joint, cascade & 53.2 & 54.2 & 11.9 & 12.6 & 3.0 & 2.9 \\
Joint, heads only & 45.4 & 45.3 & 3.1 & 4.4 & 0 & 0 \\
\bottomrule
\end{tabular}
\caption{Task success (\%) online and tokens generated per WebArena step, counting every token the generator decodes, excluding the forced prefix of a WRITE step. Cascades use $\tau=0.5$.}
\label{tab:online}
\end{table}

Table~\ref{tab:online} reports the online agents. On WebArena the cascade is the strongest system at both scales. Its heads act alone on 52\% (4B) and 51\% (8B) of the steps and decide a WRITE on another 17\% and 14\%, which brings the generated tokens below half, and episodes get shorter, from 15.7 to 14.0 steps per task for the 4B agent and from 16.8 to 16.2 for the 8B one.

\begin{table}[t]
\centering
\small
\setlength{\tabcolsep}{4pt}
\begin{tabular}{lcc}
\toprule
System & Seen template (47) & New template (127) \\
\midrule
Generator 4B & 34.8 & \textbf{4.2} \\
\ours Cascade 4B & \ours \textbf{41.8} & \ours 3.4 \\
Generator 8B & 33.3 & \textbf{6.3} \\
\ours Cascade 8B & \ours \textbf{36.9} & \ours \textbf{6.3} \\
\bottomrule
\end{tabular}
\caption{WebArena task success (\%) on the 47 test tasks whose intent template occurs among the 115 replayed training tasks, and on the other 127.}
\label{tab:watpl}
\end{table}

Table~\ref{tab:watpl} splits the test tasks by intent template. A test task counts as seen when its intent matches the template of a replayed training task, as task 513 of Figure~\ref{fig:example} matches training task 514 through ``Add a \{product\} to my wish list''. The whole gain comes from the 47 seen tasks, where the 4B cascade completes 41.8\% against 34.8\% and the 8B cascade 36.9\% against 33.3\%. On the 127 new templates both systems stay below 7\%, and the cascade trails the 4B generator by 0.8 points and ties the 8B one. On WebArena the heads therefore pay off where the replayed demonstrations cover a task's template.

On MiniWoB++ the cascade at $\tau=0.9$ reaches 53.1\% (4B) and 55.4\% (8B), within 0.2 points of the generator, with the heads acting alone on 29\% and 33\% of the steps; at $\tau=0.5$ it gives up 0.8 and 1.9 points while they act alone on 67\% and 69\%.

\subsection{Speed and the Confidence Threshold}
\label{sec:speed}

\begin{table}[t]
\centering
\small
\setlength{\tabcolsep}{4.5pt}
\begin{tabular}{lccc}
\toprule
System & s/step & Speedup & Act alone \\
\midrule
Generator & 0.234 & 1.00$\times$ & 0\% \\
\ours Cascade, $\tau=0.9$ & \ours 0.207 & \ours 1.13$\times$ & \ours 29\% \\
\ours Cascade, $\tau=0.5$ & \ours 0.168 & \ours 1.39$\times$ & \ours 67\% \\
Joint, generation & 0.238 & 1.00$\times$ & 0\% \\
Joint, cascade & 0.155 & 1.54$\times$ & 67\% \\
\bottomrule
\end{tabular}
\caption{Model time per step on MiniWoB++ (4B, one agent per GPU) and share of steps on which the heads act alone. Speedups are relative to the generation that hosts the heads: the generator for our cascades, joint generation for the joint cascade.}
\label{tab:speed}
\end{table}

Table~\ref{tab:speed} measures the model time per step on MiniWoB++ (protocol in Appendix~\ref{app:latency}). The 4B cascade is 1.13 times faster than its generator at $\tau=0.9$ and 1.39 times at $\tau=0.5$; success in these timing runs stays within 0.2 points of Section~\ref{sec:online}. The saving grows with the length of what the generator writes, which Section~\ref{sec:reason} measures for a generator that writes a thought before each action.

\begin{figure}[t]
  \centering
  \includegraphics[width=\columnwidth]{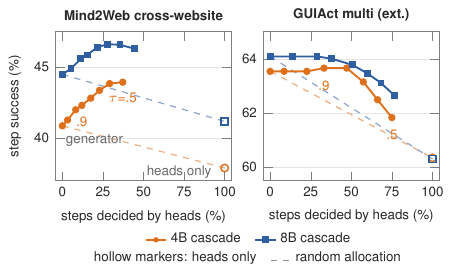}
  \caption{Step success against the share of steps the heads decide, for thresholds from 0.99 (left of each curve) to 0.5 (right). The leftmost point is the generator and the hollow marker at 100\% is heads only; dashed lines give the expected success when the same share of steps goes to the heads at random. Appendix~\ref{app:tau} lists all test sets.}
  \label{fig:tau}
\end{figure}

Figure~\ref{fig:tau} sweeps the threshold. On M2W cross-website the cascade climbs far above random allocation and peaks between $\tau=0.5$ and $\tau=0.7$, where the heads decide about a third of the steps; at $\tau=0.5$ the 4B cascade reaches 44.0\% where random allocation of the same share would give 39.8\%. On GUIAct multi (ext.) the curve stays flat down to $\tau=0.8$ and then falls toward random allocation. How low the threshold can go therefore depends on the test distribution: on M2W the cascade gains down to $\tau=0.5$, while on GUIAct it should stay at 0.8 or above.

\subsection{Comparison with Published Systems}
\label{sec:std}

Table~\ref{tab:std} in Appendix~\ref{app:std} reports M2W under the standard protocol, where every test step counts and the 86, 46 and 240 steps outside our windows count as failures. The ordering of our systems matches Table~\ref{tab:offline}: the 4B cascade improves on its generator on all three splits and the 8B cascade on cross-website and cross-domain, and the two-model cascades of Section~\ref{sec:joint} reach up to 3.8 points above the generator (48.8 against 45.0 on 8B cross-website). Published systems choose among their own candidate sets or the full page, and their rows serve as reference points. Aguvis 7B is ahead of all our systems on the three splits.

\section{Analysis}
\label{sec:analysis}

\subsection{Joint Training Costs Generation}
\label{sec:joint}

\begin{table}[t]
\centering
\small
\setlength{\tabcolsep}{5pt}
\begin{tabular}{lcccc}
\toprule
& \multicolumn{2}{c}{Generation} & \multicolumn{2}{c}{Heads only} \\
\cmidrule(lr){2-3}\cmidrule(lr){4-5}
Test set & 4B & 8B & 4B & 8B \\
\midrule
M2W task & $-4.2$ & $-4.3$ & $+1.1$ & $+1.0$ \\
M2W website & $-4.9$ & $-2.4$ & $+1.3$ & $+1.1$ \\
M2W domain & $-1.9$ & $-1.9$ & $+1.2$ & $+1.0$ \\
GUIAct single & $-0.5$ & $-0.8$ & $+1.2$ & $+1.0$ \\
GUIAct multi & $-4.0$ & $-3.9$ & $+1.6$ & $+1.7$ \\
GUIAct single, ext. & $-0.8$ & $-0.7$ & $+0.7$ & $+0.3$ \\
GUIAct multi, ext. & $-3.0$ & $-2.7$ & $+1.9$ & $+0.7$ \\
MiniWoB++ & $\phantom{+}0.0$ & $-0.5$ & $+0.9$ & $+0.6$ \\
WebArena & $-1.0$ & $-2.5$ & & \\
\bottomrule
\end{tabular}
\caption{Effect of joint training, in points of success: generation of the joint adapter minus that of the separately trained generator, and joint heads minus heads trained on the frozen generator.}
\label{tab:joint}
\end{table}

Table~\ref{tab:joint} shows what joint training costs. The joint adapter generates worse on all seven offline sets at both scales, by up to 4.9 points, and its generation loss on the validation steps rises from 0.214 to 0.235 (4B) and from 0.197 to 0.218 (8B), both for seed 0. The drop in step success is largest on the M2W cross-task and cross-website splits and on multi step GUIAct, and it carries over to WebArena.

Joint training gives stronger heads in exchange, and heads only gains 0.3 to 1.9 points. On MiniWoB++, where joint training leaves generation intact, the joint cascade reaches 53.2\% and 54.2\% against 52.5\% and 53.6\% for our cascade, and on GUIAct the two cascades at $\tau=0.5$ stay within 0.8 points of each other. On WebArena and on five of the six M2W columns our cascade is ahead, because it starts from the stronger generator.

\begin{table}[t]
\centering
\small
\setlength{\tabcolsep}{2.4pt}
\begin{tabular}{lccccc}
\toprule
& \multicolumn{3}{c}{M2W} & \multicolumn{2}{c}{GUIAct} \\
\cmidrule(lr){2-4}\cmidrule(lr){5-6}
System & Task & Web. & Dom. & Single & Multi \\
\midrule
Cascade 4B, $\tau=0.5$ & 47.6 & 44.0 & 44.6 & 82.9 & 54.2 \\
Two models 4B, $\tau=0.5$ & \textbf{47.9} & 44.3 & 46.1 & 83.6 & 56.0 \\
Two models 4B, $\tau=0.9$ & 46.7 & 41.9 & 44.5 & \textbf{84.3} & 56.0 \\
4B heads, 8B gen. & 47.8 & \textbf{45.6} & \textbf{47.2} & 82.9 & \textbf{56.5} \\
\midrule
Cascade 8B, $\tau=0.5$ & 50.2 & 46.3 & 48.2 & 82.8 & 56.9 \\
Two models 8B, $\tau=0.5$ & \textbf{51.4} & \textbf{48.5} & \textbf{49.1} & 83.7 & 57.7 \\
Two models 8B, $\tau=0.9$ & 50.7 & 45.5 & 48.3 & \textbf{84.3} & \textbf{58.5} \\
\bottomrule
\end{tabular}
\caption{Offline step success (\%) of cascades whose heads come from the joint adapter and whose generator is the separately trained one. The fourth row pairs 4B joint heads with the generation of the 8B joint adapter at $\tau=0.5$.}
\label{tab:two}
\end{table}

A two-model cascade combines both strengths: the joint heads decide, and uncertain steps go to the separately trained generator. Table~\ref{tab:two} shows that this gives the best M2W results at both scales, 51.4, 48.5 and 49.1 at 8B, at the cost of a second adapter in memory and a second prefill on every handed back step. On MiniWoB++ the 4B two-model cascade reaches 53.1\% at $\tau=0.5$ and 52.8\% at $\tau=0.9$, and pairing 4B joint heads with the 8B joint adapter reaches 54.2\%, between the two generators.

\subsection{Where Heads Beat Generation}
\label{sec:where}

\begin{table}[t]
\centering
\small
\setlength{\tabcolsep}{3.8pt}
\begin{tabular}{lcccc}
\toprule
& & \multicolumn{2}{c}{Decided steps} & Rest \\
\cmidrule(lr){3-4}\cmidrule(lr){5-5}
Test set & Share & Heads & Gen. & Gen. \\
\midrule
M2W task & 38\% & \textbf{72.2} & 68.4 & 32.7 \\
M2W website & 37\% & \textbf{68.6} & 60.3 & 29.4 \\
M2W domain & 40\% & \textbf{65.1} & 64.3 & 31.1 \\
GUIAct single & 85\% & 87.4 & \textbf{88.8} & 56.5 \\
GUIAct multi & 66\% & 67.3 & \textbf{70.2} & 28.5 \\
\midrule
\multicolumn{5}{l}{\emph{M2W website by operation} \quad Gen. $\rightarrow$ Cascade} \\
Click (781 steps) & 33\% & \multicolumn{3}{c}{45.8 $\rightarrow$ 46.0} \\
Type (120 steps) & 51\% & \multicolumn{3}{c}{24.2 $\rightarrow$ 33.3} \\
Select (72 steps) & 56\% & \multicolumn{3}{c}{14.8 $\rightarrow$ 39.8} \\
\bottomrule
\end{tabular}
\caption{The 4B cascade at $\tau=0.5$. Top: step success on the steps the heads decide, by the heads and by the generator, and on the remaining steps by the generator. Bottom: step success by operation on M2W cross-website.}
\label{tab:where}
\end{table}

Table~\ref{tab:where} splits each test set into the steps the heads decide at $\tau=0.5$ and the rest. On M2W the heads solve their steps more often than the generator would, by 8.3 points on unseen websites, and the generator solves only 29 to 33\% of the steps they leave. On GUIAct the generator is ahead on the heads' steps by 1.4 and 2.9 points; weighted by the 85\% and 66\% of steps the heads decide, this gives the 1.2 and 1.9 point loss of Table~\ref{tab:offline}.

Typing improves on every M2W split, by 9.1, 9.1 and 6.3 points (Appendix~\ref{app:tau}): the head points at goal text when it is listed and otherwise passes only the text to the generator. Selecting improves most on unseen websites, from 14.8\% to 39.8\% of 72 steps, and by 4.0 points on unseen tasks. The generator has to pair each select element with its choices in the option list and copy the text, while the argument head scores only the choices of the box it selected. Clicks change by $+0.1$, $+0.2$ and $-0.8$ points while the heads decide about a third of them.

\subsection{Target Head and Layer Weights}
\label{sec:design}

\paragraph{Box pooling against patch attention.}
\begin{table}[t]
\centering
\small
\setlength{\tabcolsep}{3pt}
\begin{tabular}{lccccc}
\toprule
& \multicolumn{3}{c}{M2W} & \multicolumn{2}{c}{GUIAct} \\
\cmidrule(lr){2-4}\cmidrule(lr){5-6}
Target head & Task & Web. & Dom. & Single & Multi \\
\midrule
Patch attention & 35.3 & 34.7 & 37.8 & \textbf{54.4} & 11.7 \\
Box pooling & \textbf{37.3} & \textbf{35.2} & \textbf{39.2} & 54.2 & \textbf{13.5} \\
\bottomrule
\end{tabular}
\caption{Heads only step success (\%) with two target heads, both trained as joint heads on MiniWoB++ and M2W; GUIAct is held out from this training.}
\label{tab:patch}
\end{table}

We replace our target head with the patch attention head of GUI-Actor \citep{wu2026gui}: the target query attends over every visual token covered by a candidate, training raises the attention mass on the tokens of the gold box, and at test time the top token selects the smallest candidate that contains it. We train both variants as joint heads on the MiniWoB++ and M2W portion of the data. Heads only with box pooling is ahead on four of five test sets (Table~\ref{tab:patch}), for instance 37.3 against 35.3 on M2W cross-task. Box pooling gives every candidate its own score. Scoring each box by the attention mass inside it instead fails on M2W, where 82\% of the test steps contain a larger candidate that encloses the gold box and always collects at least as much mass; the smallest-candidate rule of GUI-Actor is needed to avoid this.

\paragraph{Layer weights.}
The learned mix (Eq.~\ref{eq:mix}) gives the largest weight to layer 24 (0.34) and the smallest to the last layer (0.07) on the 4B generator, with the same profile on the 8B one. Under joint training the largest weight moves to layer 30 (0.38).

\subsection{Heads on a Reasoning Generator}
\label{sec:reason}

\begin{table}[t]
\centering
\small
\setlength{\tabcolsep}{4pt}
\begin{tabular}{lccc}
\toprule
System & Success & s/step & Act alone \\
\midrule
Generator, no thought & \textbf{53.9} & 0.24 & 0\% \\
Reasoning generator & 49.9 & 1.65 & 0\% \\
Joint, generation & 51.0 & 1.56 & 0\% \\
\ours Joint, cascade $\tau=0.5$ & \ours 51.5 & \ours 0.70 & \ours 66\% \\
\bottomrule
\end{tabular}
\caption{MiniWoB++ success (\%) of 4B generators that write a thought before the action. All rows are trained on MiniWoB++ only; the joint rows train the reasoning generator and the heads in one adapter.}
\label{tab:reason}
\end{table}

A generator can also write a short thought before its action, in the style of ReAct \citep{yao2022react}. We train such a 4B generator on MiniWoB++ only. Its thoughts come from sampling four thought and action pairs per step from the untuned 8B model and keeping a thought whose action matches the demonstration, in the spirit of STaR \citep{zelikman2022star}; this covers 57\% of the 7,675 deduplicated steps, and the others receive a thought written with the demonstrated action in view. Table~\ref{tab:reason} shows that thoughts do not help here: the reasoning generator trails the generator without thoughts by 4.0 points and is seven times slower.

The heads save the most on such a generator, because every step they take alone skips a whole thought. Trained in the same adapter, the joint heads act alone on two thirds of the steps, raise success from 51.0 to 51.5\% over the adapter's own generation, and cut the time per step from 1.56 to 0.70 seconds. With thoughts, the joint adapter also generates better than the separately trained reasoning generator (51.0 against 49.9).

\section{Conclusion}

A web agent can read many of its decisions from the prefill it already computes. Heads on a frozen generator decide the steps they read well and pass the rest, together with any new text, to the same generator. At a strict threshold the agent keeps the generator's accuracy, and at a looser one it gains on unseen websites and on WebArena while generating less than half of the tokens.

\section*{Limitations}

Our heads choose among numbered boxes, so agents that act through raw screen coordinates would need a different target head. The heads are tied to one generator; after fine-tuning the generator for WebArena we trained them again. All experiments use Qwen3-VL backbones and a single action grammar, and the threshold is fixed per experiment.

\bibliography{custom}

@inproceedings{shi2017world,
  title={{World of Bits}: An open-domain platform for web-based agents},
  author={Shi, Tianlin and Karpathy, Andrej and Fan, Linxi and Hernandez, Jonathan and Liang, Percy},
  booktitle={International Conference on Machine Learning},
  pages={3135--3144},
  year={2017},
  organization={PMLR}
}

@article{liu2018reinforcement,
  title={Reinforcement learning on web interfaces using workflow-guided exploration},
  author={Liu, Evan Zheran and Guu, Kelvin and Pasupat, Panupong and Shi, Tianlin and Liang, Percy},
  journal={arXiv preprint arXiv:1802.08802},
  year={2018}
}

@article{deng2023mind2web,
  title={{Mind2Web}: Towards a generalist agent for the web},
  author={Deng, Xiang and Gu, Yu and Zheng, Boyuan and Chen, Shijie and Stevens, Sam and Wang, Boshi and Sun, Huan and Su, Yu},
  journal={Advances in Neural Information Processing Systems},
  volume={36},
  pages={28091--28114},
  year={2023}
}

@inproceedings{zhou2024webarena,
  title={{WebArena}: A realistic web environment for building autonomous agents},
  author={Zhou, Shuyan and Xu, Frank F and Zhu, Hao and Zhou, Xuhui and Lo, Robert and Sridhar, Abishek and Cheng, Xianyi and Ou, Tianyue and Bisk, Yonatan and Fried, Daniel and others},
  booktitle={International Conference on Learning Representations},
  volume={2024},
  pages={15585--15606},
  year={2024}
}

@inproceedings{koh2024visualwebarena,
  title={{VisualWebArena}: Evaluating multimodal agents on realistic visual web tasks},
  author={Koh, Jing Yu and Lo, Robert and Jang, Lawrence and Duvvur, Vikram and Lim, Ming and Huang, Po-Yu and Neubig, Graham and Zhou, Shuyan and Salakhutdinov, Russ and Fried, Daniel},
  booktitle={Proceedings of the 62nd Annual Meeting of the Association for Computational Linguistics (Volume 1: Long Papers)},
  pages={881--905},
  year={2024}
}

@article{yang2023set,
  title={{Set-of-Mark} prompting unleashes extraordinary visual grounding in {GPT-4V}},
  author={Yang, Jianwei and Zhang, Hao and Li, Feng and Zou, Xueyan and Li, Chunyuan and Gao, Jianfeng},
  journal={arXiv preprint arXiv:2310.11441},
  year={2023}
}

@article{zheng2024gpt,
  title={{GPT-4V}(ision) is a generalist web agent, if grounded},
  author={Zheng, Boyuan and Gou, Boyu and Kil, Jihyung and Sun, Huan and Su, Yu},
  journal={arXiv preprint arXiv:2401.01614},
  year={2024}
}

@inproceedings{cheng2024seeclick,
  title={{SeeClick}: Harnessing {GUI} grounding for advanced visual {GUI} agents},
  author={Cheng, Kanzhi and Sun, Qiushi and Chu, Yougang and Xu, Fangzhi and YanTao, Li and Zhang, Jianbing and Wu, Zhiyong},
  booktitle={Proceedings of the 62nd Annual Meeting of the Association for Computational Linguistics (Volume 1: Long Papers)},
  pages={9313--9332},
  year={2024}
}

@inproceedings{hong2024cogagent,
  title={{CogAgent}: A visual language model for {GUI} agents},
  author={Hong, Wenyi and Wang, Weihan and Lv, Qingsong and Xu, Jiazheng and Yu, Wenmeng and Ji, Junhui and Wang, Yan and Wang, Zihan and Dong, Yuxiao and Ding, Ming and others},
  booktitle={2024 IEEE/CVF Conference on Computer Vision and Pattern Recognition (CVPR)},
  pages={14281--14290},
  year={2024},
  organization={IEEE}
}

@inproceedings{gou2025navigating,
  title={Navigating the digital world as humans do: Universal visual grounding for {GUI} agents},
  author={Gou, Boyu and Wang, Demi Ruohan and Zheng, Boyuan and Xie, Yanan and Chang, Cheng and Shu, Yiheng and Sun, Huan and Su, Yu},
  booktitle={International Conference on Learning Representations},
  volume={2025},
  pages={30851--30883},
  year={2025}
}

@inproceedings{wu2025atlas,
  title={{OS-ATLAS}: Foundation action model for generalist {GUI} agents},
  author={Wu, Zhiyong and Wu, Zhenyu and Xu, Fangzhi and Wang, Yian and Sun, Qiushi and Jia, Chengyou and Cheng, Kanzhi and Ding, Zichen and Chen, Liheng and Liang, Paul Pu and others},
  booktitle={International Conference on Learning Representations},
  volume={2025},
  pages={5090--5108},
  year={2025}
}

@article{xu2024aguvis,
  title={{Aguvis}: Unified pure vision agents for autonomous {GUI} interaction},
  author={Xu, Yiheng and Wang, Zekun and Wang, Junli and Lu, Dunjie and Xie, Tianbao and Saha, Amrita and Sahoo, Doyen and Yu, Tao and Xiong, Caiming},
  journal={arXiv preprint arXiv:2412.04454},
  year={2024}
}

@article{qin2025ui,
  title={{UI-TARS}: Pioneering automated {GUI} interaction with native agents},
  author={Qin, Yujia and Ye, Yining and Fang, Junjie and Wang, Haoming and Liang, Shihao and Tian, Shizuo and Zhang, Junda and Li, Jiahao and Li, Yunxin and Huang, Shijue and others},
  journal={arXiv preprint arXiv:2501.12326},
  year={2025}
}

@article{lu2024omniparser,
  title={{OmniParser} for pure vision based {GUI} agent},
  author={Lu, Yadong and Yang, Jianwei and Shen, Yelong and Awadallah, Ahmed},
  journal={arXiv preprint arXiv:2408.00203},
  year={2024}
}

@inproceedings{humphreys2022data,
  title={A data-driven approach for learning to control computers},
  author={Humphreys, Peter C and Raposo, David and Pohlen, Tobias and Thornton, Gregory and Chhaparia, Rachita and Muldal, Alistair and Abramson, Josh and Georgiev, Petko and Santoro, Adam and Lillicrap, Timothy},
  booktitle={International Conference on Machine Learning},
  pages={9466--9482},
  year={2022},
  organization={PMLR}
}

@article{wu2026gui,
  title={{GUI-Actor}: Coordinate-free visual grounding for {GUI} agents},
  author={Wu, Qianhui and Cheng, Kanzhi and Yang, Rui and Zhang, Chaoyun and Yang, Jianwei and Jiang, Huiqiang and Mu, Jian and Peng, Baolin and Qiao, Bo and Tan, Reuben and others},
  journal={Advances in Neural Information Processing Systems},
  volume={38},
  pages={15101--15128},
  year={2026}
}

@inproceedings{teerapittayanon2016branchynet,
  title={{BranchyNet}: Fast inference via early exiting from deep neural networks},
  author={Teerapittayanon, Surat and McDanel, Bradley and Kung, Hsiang-Tsung},
  booktitle={2016 23rd international conference on pattern recognition (ICPR)},
  pages={2464--2469},
  year={2016},
  organization={IEEE}
}

@inproceedings{xin2020deebert,
  title={{DeeBERT}: Dynamic early exiting for accelerating {BERT} inference},
  author={Xin, Ji and Tang, Raphael and Lee, Jaejun and Yu, Yaoliang and Lin, Jimmy},
  booktitle={Proceedings of the 58th annual meeting of the association for computational linguistics},
  pages={2246--2251},
  year={2020}
}

@article{schuster2022confident,
  title={Confident adaptive language modeling},
  author={Schuster, Tal and Fisch, Adam and Gupta, Jai and Dehghani, Mostafa and Bahri, Dara and Tran, Vinh and Tay, Yi and Metzler, Donald},
  journal={Advances in Neural Information Processing Systems},
  volume={35},
  pages={17456--17472},
  year={2022}
}

@article{chen2023frugalgpt,
  title={{FrugalGPT}: How to use large language models while reducing cost and improving performance},
  author={Chen, Lingjiao and Zaharia, Matei and Zou, James},
  journal={arXiv preprint arXiv:2305.05176},
  year={2023}
}

@inproceedings{leviathan2023fast,
  title={Fast inference from transformers via speculative decoding},
  author={Leviathan, Yaniv and Kalman, Matan and Matias, Yossi},
  booktitle={International conference on machine learning},
  pages={19274--19286},
  year={2023},
  organization={PMLR}
}

@article{cai2024medusa,
  title={{Medusa}: Simple {LLM} inference acceleration framework with multiple decoding heads},
  author={Cai, Tianle and Li, Yuhong and Geng, Zhengyang and Peng, Hongwu and Lee, Jason D and Chen, Deming and Dao, Tri},
  journal={arXiv preprint arXiv:2401.10774},
  year={2024}
}

@article{lin2023swiftsage,
  title={{SwiftSage}: A generative agent with fast and slow thinking for complex interactive tasks},
  author={Lin, Bill Yuchen and Fu, Yicheng and Yang, Karina and Brahman, Faeze and Huang, Shiyu and Bhagavatula, Chandra and Ammanabrolu, Prithviraj and Choi, Yejin and Ren, Xiang},
  journal={Advances in Neural Information Processing Systems},
  volume={36},
  pages={23813--23825},
  year={2023}
}

@article{bai2025qwen3,
  title={{Qwen3-VL} technical report},
  author={Bai, Shuai and Cai, Yuxuan and Chen, Ruizhe and Chen, Keqin and Chen, Xionghui and Cheng, Zesen and Deng, Lianghao and Ding, Wei and Gao, Chang and Ge, Chunjiang and others},
  journal={arXiv preprint arXiv:2511.21631},
  year={2025}
}

@article{hu2021lora,
  title={{LoRA}: Low-rank adaptation of large language models},
  author={Hu, Edward J and Shen, Yelong and Wallis, Phillip and Allen-Zhu, Zeyuan and Li, Yuanzhi and Wang, Shean and Wang, Lu and Chen, Weizhu},
  journal={arXiv preprint arXiv:2106.09685},
  year={2021}
}

@inproceedings{chen2025guicourse,
  title={{GUICourse}: From general vision language model to versatile {GUI} agent},
  author={Chen, Wentong and Cui, Junbo and Hu, Jinyi and Qin, Yujia and Fang, Junjie and Zhao, Yue and Wang, Chongyi and Liu, Jun and Chen, Guirong and Huo, Yupeng and others},
  booktitle={Proceedings of the 63rd Annual Meeting of the Association for Computational Linguistics (Volume 1: Long Papers)},
  pages={21936--21959},
  year={2025}
}

@article{lu2024weblinx,
  title={{WebLINX}: Real-world website navigation with multi-turn dialogue},
  author={L{\`u}, Xing Han and Kasner, Zden{\v{e}}k and Reddy, Siva},
  journal={arXiv preprint arXiv:2402.05930},
  year={2024}
}

@article{zelikman2022star,
  title={{STaR}: Bootstrapping reasoning with reasoning},
  author={Zelikman, Eric and Wu, Yuhuai and Mu, Jesse and Goodman, Noah},
  journal={Advances in Neural Information Processing Systems},
  volume={35},
  pages={15476--15488},
  year={2022}
}

@article{chezelles2024browsergym,
  title={The {BrowserGym} ecosystem for web agent research},
  author={Chezelles, De and Le Sellier, Thibault and Shayegan, Sahar Omidi and Jang, Lawrence Keunho and L{\`u}, Xing Han and Yoran, Ori and Kong, Dehan and Xu, Frank F and Reddy, Siva and Cappart, Quentin and others},
  journal={arXiv preprint arXiv:2412.05467},
  year={2024}
}

@inproceedings{peters-etal-2018-deep,
    title = "Deep Contextualized Word Representations",
    author = "Peters, Matthew E.  and
      Neumann, Mark  and
      Iyyer, Mohit  and
      Gardner, Matt  and
      Clark, Christopher  and
      Lee, Kenton  and
      Zettlemoyer, Luke",
    editor = "Walker, Marilyn  and
      Ji, Heng  and
      Stent, Amanda",
    booktitle = "Proceedings of the 2018 Conference of the {N}orth {A}merican Chapter of the {A}ssociation for {C}omputational {L}inguistics: Human Language Technologies, Volume 1 (Long Papers)",
    month = jun,
    year = "2018",
    address = "New Orleans, Louisiana",
    publisher = "Association for Computational Linguistics",
    url = "https://aclanthology.org/N18-1202/",
    doi = "10.18653/v1/N18-1202",
    pages = "2227--2237"
}

@article{alain2016understanding,
  title={Understanding intermediate layers using linear classifier probes},
  author={Alain, Guillaume and Bengio, Yoshua},
  journal={arXiv preprint arXiv:1610.01644},
  year={2016}
}

@article{koh2024tree,
  title={Tree search for language model agents},
  author={Koh, Jing Yu and McAleer, Stephen and Fried, Daniel and Salakhutdinov, Ruslan},
  journal={arXiv preprint arXiv:2407.01476},
  year={2024}
}

@article{loshchilov2017decoupled,
  title={Decoupled weight decay regularization},
  author={Loshchilov, Ilya and Hutter, Frank},
  journal={arXiv preprint arXiv:1711.05101},
  year={2017}
}

@article{yao2022react,
  title={{ReAct}: Synergizing reasoning and acting in language models},
  author={Yao, Shunyu and Zhao, Jeffrey and Yu, Dian and Du, Nan and Shafran, Izhak and Narasimhan, Karthik and Cao, Yuan},
  journal={arXiv preprint arXiv:2210.03629},
  year={2022}
}

@article{vinyals2015pointer,
  title={Pointer networks},
  author={Vinyals, Oriol and Fortunato, Meire and Jaitly, Navdeep},
  journal={Advances in neural information processing systems},
  volume={28},
  year={2015}
}

@article{geifman2017selective,
  title={Selective classification for deep neural networks},
  author={Geifman, Yonatan and El-Yaniv, Ran},
  journal={Advances in neural information processing systems},
  volume={30},
  year={2017}
}

@article{jiang2025kore,
  title={{KORE}: Enhancing Knowledge Injection for Large Multimodal Models via Knowledge-Oriented Augmentations and Constraints},
  author={Jiang, Kailin and Jiang, Hongbo and Jiang, Ning and Gao, Zhi and Bi, Jinhe and Ren, Yuchen and Li, Bin and Du, Yuntao and Liu, Lei and Li, Qing},
  journal={arXiv preprint arXiv:2510.19316},
  year={2025}
}

@inproceedings{jiang2026mined,
  title={{MINED}: Probing and updating with multimodal time-sensitive knowledge for large multimodal models},
  author={Jiang, Kailin and Jiang, Ning and Du, Yuntao and Ren, Yuchen and Li, Yuchen and Gao, Yifan and Bi, Jinhe and Ma, Yunpu and Li, Bin and Liu, Lei and others},
  booktitle={Findings of the Association for Computational Linguistics: ACL 2026},
  pages={13766--13795},
  year={2026}
}

@article{peng2025can,
  title={Can Visual Input Be Compressed? A Visual Token Compression Benchmark for Large Multimodal Models},
  author={Peng, Tianfan and Du, Yuntao and Ji, Pengzhou and Dong, Shijie and Jiang, Kailin and Ma, Mingchuan and Tian, Yijun and Bi, Jinhe and Li, Qian and Du, Wei and others},
  journal={arXiv preprint arXiv:2511.02650},
  year={2025}
}

@inproceedings{jiang2026large,
  title={When large multimodal models confront evolving knowledge: Challenges and explorations},
  author={Jiang, Kailin and Du, Yuntao and Ding, Yukai and Ren, Yuchen and Jiang, Ning and Gao, Zhi and Zheng, Zilong and Liu, Lei and Li, Bin and Li, Qing},
  booktitle={The Fourteenth International Conference on Learning Representations},
  year={2026}
}

@inproceedings{jia2026benchmarking,
  title={Benchmarking multimodal knowledge conflict for large multimodal models},
  author={Jia, Yifan and Du, Yuntao and Jiang, Kailin and Liang, Yuyang and Ren, Qihan and Xin, Yi and Yang, Rui and Feng, Fenze and Chen, MingCai and Lu, Hengyang and others},
  booktitle={Proceedings of the AAAI Conference on Artificial Intelligence},
  volume={40},
  number={27},
  pages={22283--22291},
  year={2026}
}

@inproceedings{jiang2025mmke,
  title={{MMKE-Bench}: A multimodal editing benchmark for diverse visual knowledge},
  author={Jiang, Kailin and Gao, Zhi and Shi, Chenrui and Zheng, Zilong and Qi, Siyuan and Li, Qing and others},
  booktitle={International Conference on Learning Representations},
  volume={2025},
  pages={526--555},
  year={2025}
}

@inproceedings{qi2025context,
  title={In-context editing: Learning knowledge from self-induced distributions},
  author={Qi, Siyuan and Yang, Bangcheng and Jiang, Kailin and Wang, Xiaobo and Li, Jiaqi and Zhong, Yifan and Yang, Yaodong and Zheng, Zilong},
  booktitle={International Conference on Learning Representations},
  volume={2025},
  pages={77563--77585},
  year={2025}
}

@article{jiang2026beyond,
  title={Beyond relevance-centric retrieval: Rubric-oriented document set selection and ranking},
  author={Jiang, Kailin and Liu, Lei and Xi, Jian and Xu, Hui and Liu, Junlin and Fu, Baochen and Li, Bin and Lu, Yu and Shi, Haibo and others},
  journal={arXiv preprint arXiv:2607.19747},
  year={2026}
}

@article{liu2026proprietary,
  title={From proprietary to open-source: Bridging the distribution gap via multi-agent protocol distillation in agentic search},
  author={Liu, Junlin and Chen, Jiangwang and Song, Zixin and Zhou, Shuaiyu and Lv, Chunji and Wu, Hank and Jiang, Kailin and Wu, Jinyang and Yu, Bohan and Zhou, Chenxi},
  journal={arXiv preprint arXiv:2607.24280},
  year={2026}
}

@article{fu2026can,
  title={Can Multimodal Large Language Models Understand {OCT}?},
  author={Fu, Baochen and Deng, Wenzhi and Jin, Baihao and Li, Yang and Nie, Zihan and Jiang, Kailin and Du, Yuntao and Song, Weiye},
  journal={arXiv preprint arXiv:2607.16609},
  year={2026}
}

@article{jiang2026adatutorank,
  title={{AdaTutoRank}: Learning to Rerank Document Sets via Adaptive Tutoring Optimization for {RAG} and Deep Research},
  author={Jiang, Kailin and Liu, Lei and Xi, Jian and Chen, Yangqi and Xu, Hui and Zhao, Hongwei and Li, Bin and Lu, Yu and Shi, Haibo},
  journal={arXiv preprint arXiv:2609.32472},
  year={2026}
}

@article{li2026knowhal,
  title={{KnowHal}: A Knowledge-Driven Benchmark for Comprehensive Multimodal Hallucination Evaluation},
  author={Li, Ruihan and Tan, Jiyang and Jiang, Kailin and Li, Huining and Lu, Hengyang and Huang, Yu and Li, Qian and Du, Yuntao},
  journal={arXiv preprint arXiv:2608.03782},
  year={2026}
}

\appendix
\onecolumn

\section{Data Conversion}
\label{app:data}

\begin{table}[!ht]
\centering
\small
\setlength{\tabcolsep}{4pt}
\begin{tabular}{lrl}
\toprule
Source & Steps & Main operations \\
\midrule
MiniWoB++ & 9,171 & click, type, no op \\
Mind2Web train & 7,320 & click, type, select \\
GUIAct single & 66,089 & click, scroll, answer \\
GUIAct multi & 27,700 & click, scroll, answer \\
UGround & 65,066 & click \\
WebLINX & 38,171 & click, type, scroll \\
\midrule
Total & 213,517 & \\
\midrule
WebArena replay & 1,298 & click, type, answer \\
\bottomrule
\end{tabular}
\caption{Converted training steps per source, before deduplication. MiniWoB++ excludes the held-out tasks; the WebArena replay is used only for the WebArena fine-tuning.}
\label{tab:data}
\end{table}

Every source is converted into the same step record: a screenshot, the candidate boxes in reading order, the element list with roles and names, the goal, the history and the demonstrated action in our grammar (Table~\ref{tab:data}). Boxes are drawn at load time with a colored frame and the number inside the top left corner, so that the label never covers a neighboring element. Web pages are rendered 960 pixels wide. After removing 4,472 duplicate steps and 426 steps whose action cannot be parsed, 208,619 steps remain, 7,675 of them from MiniWoB++. We hold out 10,430 steps for validation and repeat the remaining MiniWoB++ steps three times, which gives 212,761 training examples.

\paragraph{MiniWoB++.}
Episodes come from successful rollouts of earlier text based policies and scripted demonstrations on training seeds, with a screenshot saved at every step. The task area is cropped and enlarged by a factor of two. Twelve tasks from ten families (for example \texttt{bisect-angle}, \texttt{enter-password}, \texttt{use-colorwheel}) are removed from training and kept for evaluation.

\paragraph{Mind2Web.}
The window of Section~\ref{sec:bench} is scaled by 0.75. Negative containers that cover more than a quarter of the window and boxes smaller than 6 pixels are dropped. Hover and enter count as click, as in the original benchmark. The history uses the dataset's own text form of previous actions, and all positive candidates inside the window are gold boxes.

\paragraph{GUIAct.}
Each screenshot ships with its element list, which gives the candidates. Click, hover, input and select become our actions, and the extended versions keep scroll and answer as well. Multi step rows are grouped into episodes by record.

\paragraph{UGround and WebLINX.}
UGround screenshots carry many referring expressions with boxes; every box becomes a candidate, and two expressions per screenshot (32,533 screenshots) become goals of the form ``Click the element: \dots''. WebLINX demonstrations give clicks, text input and selections on real websites; the goal is the instructor's last two utterances, and candidates are the visible elements of the page's box file.

\paragraph{Text options.}
The options listed after the image are, in order: each quoted span of the goal in its original, upper, lower and capitalized form; when the goal has no quotes, each capitalized word after the first and each number in the goal, together with the names (at most three words and 25 characters) of page elements whose role is cell, text, label or list item, up to 16 strings; then the choices of every select element. The list is cut at 24 lines, each written as ``\texttt{- text .}''. In 7,641 training steps the demonstrated text is not among the options.

\FloatBarrier
\section{Training Details}
\label{app:train}

\begin{table}[!ht]
\centering
\footnotesize
\setlength{\tabcolsep}{4pt}
\begin{tabular}{ll}
\toprule
Setting & Value \\
\midrule
LoRA rank, scale, dropout & 16, 32, 0.05 \\
LoRA targets & attention and MLP \\
LoRA parameters (4B / 8B) & 33.0M / 43.6M \\
Head parameters (4B / 8B) & 6.0M / 9.2M \\
Head width $d_h$ & 512 \\
Layers read & 12, 18, 24, 30, 36 \\
Optimizer & AdamW, weight decay 0 \\
LR (LoRA / heads) & $10^{-4}$ / $10^{-3}$ \\
Schedule & 5\% warmup, cosine \\
Gradient clipping & 1.0 \\
Examples per update & 16 \\
Epochs & 1 \\
Train / val. examples & 212,761 / 10,430 \\
WebArena tuning & 3 epochs, LR $\frac{1}{2}$ of the above \\
Decoding & greedy, at most 24 tokens \\
GPU & RTX PRO 6000, 96 GB \\
\bottomrule
\end{tabular}
\caption{Training settings shared by all runs. Generators that write a thought (Section~\ref{sec:reason}) train for three epochs on MiniWoB++ only and decode at most 160 tokens.}
\label{tab:hyper}
\end{table}

Table~\ref{tab:hyper} lists the settings. We optimize with AdamW \citep{loshchilov2017decoupled}. The generator and the joint adapter train with data parallelism on two GPUs; the 4B joint adapter takes about eleven hours. Heads on a frozen generator train with the backbone in inference mode, so only the heads keep optimizer state. On the validation steps of seed 0, the frozen 4B heads predict the type correctly 92.0\% of the time, the target box 74.6\% and the argument 86.1\%; the 8B heads reach 92.2\%, 75.2\% and 87.2\%. The 8B heads weight layers 12 to 36 by 0.24, 0.25, 0.30, 0.14 and 0.08, and the joint 4B heads by 0.14, 0.13, 0.21, 0.38 and 0.14.

\FloatBarrier
\section{WebArena Harness and Expert Replay}
\label{app:wa}

\paragraph{Split and replay.}
We use 174 test tasks of WebArena (gitlab 44, shopping admin 39, shopping 34, reddit 29, map 24, wikipedia 4) and a disjoint set of training tasks, split at the task level. Expert trajectories come from a GPT-5.5 Set-of-Mark agent on the training tasks; we keep its 234 successful trajectories on 139 tasks. Because our boxes differ from the expert's, we replay each trajectory in the live sites, record at every step what our agent would see (the viewport scaled from $1440\times900$ to $960\times600$, our candidate boxes and the history), and write the expert action in our grammar. A step whose target is not among our boxes is executed but not kept, and a trajectory is dropped when the page diverges or the replay is not scored successful again. This yields 191 trajectories on 115 tasks with 1,298 steps; after removing 369 duplicates and 13 unparsable steps, 871 steps train and 45 validate.

\paragraph{Fine-tuning and evaluation.}
The joint adapter is fine-tuned on $\mathcal{L}_{\mathrm{gen}}+\mathcal{L}_{\mathrm{heads}}$ with the same schedule. Each run resets the six sites, logs in again, and runs one worker per site copy with three workers per GPU. Actions run on the live page: type replaces a field's content, enter presses Enter on a field, answer ends the episode with the text. The generator decodes at most 24 tokens per action, and an action that does not parse is an invalid step; on the WRITE path the text is cut at the first quote and at 60 characters. Tasks that hit an environment error count as failures for every system.

\begin{table}[!ht]
\centering
\small
\setlength{\tabcolsep}{2.7pt}
\begin{tabular}{lcccccc}
\toprule
System & Shop & Admin & Reddit & GitLab & Map & Wiki \\
\midrule
Generator 4B & 11.8 & \textbf{8.5} & 20.7 & 12.1 & \textbf{12.5} & \textbf{0.0} \\
\ours Cascade 4B & \ours \textbf{17.6} & \ours 4.3 & \ours \textbf{23.0} & \ours \textbf{15.2} & \ours \textbf{12.5} & \ours \textbf{0.0} \\
Generator 8B & 10.8 & \textbf{14.5} & 17.2 & \textbf{16.7} & 8.3 & \textbf{0.0} \\
\ours Cascade 8B & \ours \textbf{18.6} & \ours 11.1 & \ours \textbf{19.5} & \ours 14.4 & \ours \textbf{11.1} & \ours \textbf{0.0} \\
\bottomrule
\end{tabular}
\caption{WebArena task success (\%) per site.}
\label{tab:wasite}
\end{table}

Table~\ref{tab:wasite} breaks the results down by site. The cascade gains most on the shopping site at both scales, by 5.8 (4B) and 7.8 (8B) points, and loses on the shopping admin site. At $\tau=0.9$ the joint 4B cascade completes 11.9\% of the tasks, the same as at $\tau=0.5$, while its heads act alone on 12\% of the steps, against 51\% at $\tau=0.5$.

\FloatBarrier
\section{Per-Seed Results}
\label{app:seeds}

\begin{table}[!ht]
\centering
\footnotesize
\setlength{\tabcolsep}{1.8pt}
\begin{tabular}{llccccccc}
\toprule
& System & M2W task & M2W website & M2W domain & GUIAct single & GUIAct multi & Single, ext. & Multi, ext. \\
\midrule
\multirow{4}{*}{4B} & Generator & 45.8/46.3/46.4 & 40.2/40.7/41.7 & 44.1/44.4/44.5 & 84.8/84.0/83.4 & 54.4/56.8/57.0 & 79.4/79.0/78.4 & 62.8/64.1/63.8 \\
& Cascade 0.9 & 46.1/46.3/47.2 & 41.6/42.2/43.1 & 44.3/44.5/44.5 & 84.8/83.8/83.4 & 54.4/56.8/56.5 & 79.4/78.8/78.2 & 63.0/64.3/63.7 \\
& Cascade 0.5 & 46.9/47.9/48.0 & 43.9/44.2/43.8 & 44.5/44.9/44.5 & 84.0/82.3/82.4 & 53.4/54.4/54.6 & 78.7/77.4/77.4 & 61.6/62.5/61.4 \\
& Heads only & 39.3/40.0/39.7 & 38.1/37.8/37.7 & 39.0/39.3/39.0 & 82.6/80.0/80.4 & 52.7/56.5/54.9 & 77.0/75.0/75.5 & 60.0/61.7/59.4 \\
\midrule
\multirow{4}{*}{8B} & Generator & 49.6/51.4/50.1 & 45.1/44.7/43.7 & 47.8/48.1/48.3 & 84.7/84.8/83.7 & 58.7/59.6/58.2 & 79.1/79.7/78.9 & 63.5/64.2/64.6 \\
& Cascade 0.9 & 50.0/51.4/50.4 & 46.5/46.2/44.9 & 47.7/48.3/48.3 & 84.7/84.5/83.3 & 58.7/59.6/57.7 & 79.3/79.3/78.5 & 63.6/64.2/64.3 \\
& Cascade 0.5 & 49.6/50.8/50.1 & 46.1/46.8/46.0 & 47.7/48.5/48.4 & 82.9/83.5/82.1 & 56.8/57.7/56.3 & 78.1/78.5/77.4 & 61.9/62.8/63.2 \\
& Heads only & 42.3/41.6/41.3 & 40.3/42.1/41.1 & 41.8/41.6/41.7 & 81.5/81.6/80.7 & 55.3/55.3/54.2 & 76.4/76.5/75.5 & 60.0/60.3/60.7 \\
\bottomrule
\end{tabular}
\caption{Offline step success (\%) for seeds 0 / 1 / 2.}
\label{tab:seeds_off}
\end{table}

\begin{table}[!ht]
\centering
\small
\setlength{\tabcolsep}{2.6pt}
\begin{tabular}{lcc}
\toprule
System & 4B & 8B \\
\midrule
\multicolumn{3}{l}{\emph{MiniWoB++}} \\
Generator & 54.1 / 53.3 / 52.4 & 56.1 / 55.5 / 55.0 \\
Cascade 0.9 & 54.1 / 53.1 / 52.1 & 55.6 / 55.6 / 54.8 \\
Cascade 0.5 & 52.9 / 53.4 / 51.3 & 53.1 / 54.3 / 53.4 \\
Heads only & 44.9 / 44.9 / 43.9 & 44.4 / 44.4 / 45.4 \\
\midrule
\multicolumn{3}{l}{\emph{WebArena}} \\
Generator & 12.1 / 11.5 / 13.8 & 13.8 / 14.9 / 12.1 \\
Cascade 0.5 & 13.2 / 13.8 / 14.4 & 14.9 / 15.5 / 13.2 \\
\bottomrule
\end{tabular}
\caption{Online task success (\%) for seeds 0 / 1 / 2.}
\label{tab:seeds_on}
\end{table}

Tables~\ref{tab:seeds_off} and~\ref{tab:seeds_on} list every seed of the main systems. A seed fixes the training run: the generator of seed $s$ hosts the heads of seed $s$, and both online benchmarks use the same environment seeds and tasks for all systems.

\FloatBarrier
\section{Threshold Curves and Operations}
\label{app:tau}

\begin{table}[!ht]
\centering
\small
\setlength{\tabcolsep}{4.2pt}
\begin{tabular}{llcccccccc}
\toprule
& Test set & $\tau=0.5$ & 0.6 & 0.7 & 0.8 & 0.9 & 0.95 & 0.99 & Generator \\
\midrule
\multirow{7}{*}{4B} & M2W task & \textbf{47.6} (38) & \textbf{47.6} (30) & 47.2 (23) & 47.0 (18) & 46.6 (13) & 46.4 (9) & 46.2 (5) & 46.2 \\
& M2W website & \textbf{44.0} (37) & 43.9 (29) & 43.4 (23) & 42.8 (18) & 42.3 (12) & 42.0 (8) & 41.3 (3) & 40.9 \\
& M2W domain & 44.6 (40) & \textbf{44.7} (32) & 44.6 (25) & 44.5 (19) & 44.4 (14) & 44.3 (10) & 44.3 (5) & 44.3 \\
& GUIAct single & 82.9 (85) & 83.2 (79) & 83.6 (73) & 83.8 (66) & 84.0 (56) & \textbf{84.1} (47) & \textbf{84.1} (25) & \textbf{84.1} \\
& GUIAct multi & 54.2 (66) & 54.7 (57) & 55.3 (48) & 55.8 (39) & 55.9 (27) & 55.9 (19) & \textbf{56.1} (8) & \textbf{56.1} \\
& Single, ext. & 77.9 (87) & 78.2 (82) & 78.5 (75) & 78.7 (69) & 78.8 (60) & \textbf{78.9} (51) & \textbf{78.9} (33) & \textbf{78.9} \\
& Multi, ext. & 61.8 (75) & 62.5 (66) & 63.2 (57) & \textbf{63.7} (47) & \textbf{63.7} (33) & 63.5 (23) & 63.5 (9) & 63.5 \\
\midrule
\multirow{7}{*}{8B} & M2W task & 50.2 (46) & 50.7 (38) & 50.7 (30) & \textbf{50.8} (24) & 50.6 (17) & 50.6 (13) & 50.4 (7) & 50.4 \\
& M2W website & 46.3 (44) & \textbf{46.6} (35) & \textbf{46.6} (28) & 46.4 (22) & 45.9 (15) & 45.6 (11) & 44.9 (5) & 44.5 \\
& M2W domain & 48.2 (46) & \textbf{48.6} (38) & 48.5 (31) & 48.4 (24) & 48.1 (18) & 48.0 (13) & 48.0 (7) & 48.1 \\
& GUIAct single & 82.8 (88) & 83.3 (82) & 83.7 (77) & 84.0 (70) & 84.1 (61) & 84.3 (53) & \textbf{84.4} (34) & \textbf{84.4} \\
& GUIAct multi & 56.9 (68) & 57.6 (60) & 58.0 (51) & 58.4 (43) & 58.7 (33) & \textbf{58.9} (26) & 58.8 (12) & 58.8 \\
& Single, ext. & 78.0 (89) & 78.3 (84) & 78.7 (79) & 78.9 (72) & 79.0 (64) & 79.1 (57) & 79.2 (41) & \textbf{79.3} \\
& Multi, ext. & 62.6 (77) & 63.1 (68) & 63.5 (60) & 63.8 (51) & 64.0 (38) & \textbf{64.1} (29) & \textbf{64.1} (14) & \textbf{64.1} \\
\bottomrule
\end{tabular}
\caption{Step success (\%) of the cascade across thresholds, with the share of steps the heads decide in parentheses. For every test step we store the generator's action and, when $c_t\ge0.5$, the heads' action (with the generator's text for WRITE), so one run yields every $\tau\ge0.5$.}
\label{tab:tau}
\end{table}

\begin{table}[!ht]
\centering
\small
\setlength{\tabcolsep}{3.4pt}
\begin{tabular}{llcccc}
\toprule
Split & Op. & Steps & Gen. & Heads & Cascade \\
\midrule
\multirow{3}{*}{Task} & Click & 1,037 & 48.5 & 43.4 & 48.6 \\
& Type & 157 & 36.1 & 19.5 & 45.2 \\
& Select & 59 & 32.2 & 28.2 & 36.2 \\
\midrule
\multirow{3}{*}{Website} & Click & 781 & 45.8 & 40.8 & 46.0 \\
& Type & 120 & 24.2 & 17.5 & 33.3 \\
& Select & 72 & 14.8 & 40.3 & 39.8 \\
\midrule
\multirow{3}{*}{Domain} & Click & 3,172 & 48.5 & 43.3 & 47.7 \\
& Type & 553 & 22.6 & 16.0 & 28.9 \\
& Select & 95 & 33.7 & 32.3 & 33.0 \\
\bottomrule
\end{tabular}
\caption{Step success (\%) by operation on M2W for the 4B generator, heads only, and the cascade at $\tau=0.5$.}
\label{tab:ops}
\end{table}

Table~\ref{tab:tau} gives the curves of Figure~\ref{fig:tau} for every test set at both scales, and Table~\ref{tab:ops} the operation breakdown on all M2W splits. Heads only is weak on typing because of its WRITE fallback (Section~\ref{sec:systems}). On GUIAct, hover steps are one source of the cascade's loss, since the type mask allows hover only on menu items, which GUIAct elements never are; the cascade solves 5.4\% of the 49 hover steps of the multi step test against 10.9\% for the generator.

\FloatBarrier
\section{Latency Measurement}
\label{app:latency}

Timing runs execute one agent process per GPU with batch size one in bfloat16, with the LoRA merged into the backbone, and time each step with Hugging Face \texttt{generate} from building the prompt, including image preprocessing and the vision encoder, to the parsed action; we average over every step of the 595 MiniWoB++ episodes. The heads add attention pooling and a few projections after the prefill. In WebArena, three agents share a GPU with the browser, and the model time per step falls from 0.414 to 0.364 seconds with the 4B cascade and from 0.369 to 0.240 with the 8B one. Table~\ref{tab:online} reports generated tokens as the WebArena cost, since they do not depend on GPU sharing.

\FloatBarrier
\section{Mind2Web Standard Protocol}
\label{app:std}

\begin{table}[!ht]
\centering
\small
\setlength{\tabcolsep}{4.6pt}
\begin{tabular}{lccccccccc}
\toprule
& \multicolumn{3}{c}{Cross-task} & \multicolumn{3}{c}{Cross-website} & \multicolumn{3}{c}{Cross-domain} \\
\cmidrule(lr){2-4}\cmidrule(lr){5-7}\cmidrule(lr){8-10}
System & Ele.Acc & Op.F1 & Step SR & Ele.Acc & Op.F1 & Step SR & Ele.Acc & Op.F1 & Step SR \\
\midrule
MindAct, Flan-T5 XL (HTML) & & & 52.0 & & & 38.9 & & & 39.6 \\
SeeClick 9.6B & & & 25.5 & & & 16.4 & & & 20.8 \\
SeeAct, GPT-4V & & & 40.2 & & & 32.4 & & & 36.8 \\
OmniParser + GPT-4V & & & 39.4 & & & 36.5 & & & 42.0 \\
Aguvis 7B & & & 60.4 & & & 54.6 & & & 56.6 \\
\midrule
Generator 4B & 49.4 & 84.6 & 45.5 & 44.6 & 84.0 & 41.1 & 48.5 & 81.1 & 42.1 \\
\ours Cascade 4B, $\tau=0.5$ & \ours 51.0 & \ours 85.0 & \ours 46.6 & \ours 47.7 & \ours 84.8 & \ours 43.1 & \ours 49.0 & \ours 81.4 & \ours 42.4 \\
Two models 4B, $\tau=0.5$ & 51.5 & 84.9 & 46.8 & 48.5 & 85.1 & 43.9 & 49.9 & 81.8 & 43.7 \\
Generator 8B & 54.3 & 84.8 & 49.3 & 48.6 & 84.3 & 45.0 & 51.8 & 81.2 & 45.6 \\
\ours Cascade 8B, $\tau=0.5$ & \ours 54.7 & \ours 85.0 & \ours 49.2 & \ours 50.9 & \ours 85.0 & \ours 46.3 & \ours 51.9 & \ours 81.4 & \ours 45.7 \\
Two models 8B, $\tau=0.5$ & 55.6 & 85.0 & 50.1 & 53.5 & 86.2 & 48.8 & 52.3 & 81.7 & 46.4 \\
\bottomrule
\end{tabular}
\caption{Multimodal-Mind2Web under the standard protocol: element accuracy, operation F1 and step success rate, averaged within tasks and then over tasks, with every test step counted and steps whose gold element lies outside our window counted as failures. Published numbers are copied from MindAct \citep{deng2023mind2web}, SeeClick \citep{cheng2024seeclick}, SeeAct \citep{zheng2024gpt}, OmniParser \citep{lu2024omniparser} and Aguvis \citep{xu2024aguvis}, which use their own candidate sets or full pages; our rows use the windows and candidate lists of Section~\ref{sec:bench}.}
\label{tab:std}
\end{table}

\FloatBarrier
\section{Further Related Work}
\label{app:related}

\paragraph{Knowledge inside large multimodal models.}
Our heads read a decision that a large multimodal model has already formed in its hidden states. A neighboring line of work studies the knowledge these models hold and how it can be measured and changed. MINED probes and updates time-sensitive multimodal knowledge \citep{jiang2026mined}, \citet{jiang2026large} examine how such models cope with evolving knowledge, and KORE injects new knowledge through knowledge-oriented augmentations and constraints \citep{jiang2025kore}. MMKE-Bench evaluates the editing of diverse visual knowledge \citep{jiang2025mmke}, in-context editing learns knowledge from self-induced distributions \citep{qi2025context}, and \citet{jia2026benchmarking} benchmark conflicts between multimodal knowledge sources. KnowHal measures knowledge-driven hallucination \citep{li2026knowhal}, and \citet{fu2026can} test whether multimodal language models understand OCT images. These studies treat what the model knows as the object to probe or edit; our heads treat what the model has decided as the quantity to read out.

\paragraph{Ranking, distillation and efficient inference.}
Choosing a box among candidates is a ranking problem, and the same holds for choosing documents in retrieval augmented generation and deep research, where rubric-oriented selection \citep{jiang2026beyond} and adaptive tutoring \citep{jiang2026adatutorank} rank whole document sets. Our WebArena training data distill the trajectories of a proprietary agent into an open model, close in spirit to multi-agent protocol distillation for agentic search \citep{liu2026proprietary}. Visual token compression \citep{peng2025can} lowers the cost of the multimodal prefill that the cascade reuses, so its saving adds to the decoding that the heads skip.

\FloatBarrier
\section{Case Studies}
\label{app:cases}

The three episodes below come from WebArena test tasks whose intent template does not occur among the replayed training tasks.

\paragraph{Reddit, task 402.}
For ``Change my reddit bio to `Freelance Web Developer','' the 4B heads decide all six steps of seed 0: three navigation clicks with confidence above 0.95, typing the bio, where the argument head points at the quoted text that the goal provides (0.9998), a final click (0.86), and the answer. The episode generates only the final answer.

\paragraph{Shopping admin, task 14.}
For ``Tell me the number of reviews that our store received by far that mention term `not useful','' the 4B cascade of seed 1 hands the first click to the generator (0.49), lets the heads take the next click (0.985) and type the quoted term (0.873), hands another click to the generator (0.34), and answers \act{0} through WRITE.

\paragraph{Shopping admin, task 773.}
For ``Delete all pending negative reviews'' the 8B heads of seed 0 act alone on all fourteen clicks, with confidence between 0.66 and 0.98: three navigation clicks and then a three-click sequence on boxes 11 (9 the first time), 12 and 44 that repeats for each review, before they answer through WRITE.

\FloatBarrier
\end{document}